\documentclass[twoside,11pt]{article}

\usepackage{jmlr2e}

\usepackage{graphicx}
\usepackage{amsmath,amsfonts,amssymb,bm,eucal,wrapfig}
\usepackage{xspace}
\usepackage{url}

\firstpageno{1}

\def\tr{^\top}
\def\a{\alpha}
\def\l{\lambda}
\def\g{\gamma}
\def\la{($\lambda$)\xspace}
\def\thet{\theta}

\def\r{\rho}

\def\rr{{\bf r}}
\def\th{{\bm \theta}}
\def\ph{{\bm \phi}}
\def\ee{{\bf e}}

\def\lt{\left}
\def\rt{\right}
\def\ra{\rightarrow}

\def\I{{\bf I}}
\def\P{{\bf P}}
\def\Ppi{\P_\pi}
\def\SS{{\cal S}}
\def\AA{{\cal A}}
\def\vpi{v_\pi}

\def\Re{\mathbb R}
\def\CP{\mathbb P}
\def\CEpi#1#2{{\mathbb E}_\pi\!\left[#1\middle\vert{#2}\right]}
\def\CP#1#2{{\mathbb P}\!\left\{#1\middle\vert{#2}\right\}}
\def\Epi#1{{\mathbb E}_\pi\!\left[#1\right]}

\newtheorem{assumption}{Assumption}
\def\Gm{{\bm \Gamma}}
\def\Lm{{\bm \Lambda}}
\def\Ph{{\bm \Phi}}
\def\M{{\bf M}}
\def\m{m}
\def\A{{\bf A}}
\def\b{{\bf b}}
\def\Dmi{{{\bf d}_{\mu,i}}}
\def\dmi{d_{\mu,i}}
\def\dm{d_\mu}
\def\Vpi{{\bm v}_\pi}
\def\vv{{\bm v}}

\def\ETD{ETD}

\begin{document}

\title{Emphatic Temporal-Difference Learning}

\author{ \vspace{-0.75cm}
\center \name {A. Rupam Mahmood~~~~ Huizhen Yu~~~~ Martha White~~~~ Richard S. Sutton }\\
\addr Reinforcement Learning and Artificial Intelligence Laboratory \\
Department of Computing Science, University of Alberta, Edmonton, AB T6G 2E8 Canada \\[0.5cm]
}


\maketitle

\begin{abstract}

Emphatic algorithms are temporal-difference learning algorithms that change their effective state distribution by selectively emphasizing and de-emphasizing their updates on different time steps. Recent works by Sutton, Mahmood and White (2015), and Yu (2015) show that by varying the emphasis in a particular way, these algorithms become stable and convergent under off-policy training with linear function approximation. This paper serves as a unified summary of the available results from both works. In addition, we demonstrate the empirical benefits from the flexibility of emphatic algorithms, including state-dependent discounting, state-dependent bootstrapping, and the user-specified allocation of function approximation resources.

\end{abstract}


\section{Introduction}

A fundamental problem in reinforcement learning involves learning a sequence of long-term predictions in a dynamical system. This problem is often formulated as learning approximations to value functions of Markov decision processes (Bertsekas \& Tsitsiklis 1996, Sutton \& Barto 1998). Temporal-difference learning algorithms, such as TD\la (Sutton 1988), GQ\la (Maei \& Sutton 2010), and LSTD\la (Boyan 1999, Bradtke \& Barto 1996), 
provide effective solutions to this problem. These algorithms stand out particularly because of their ability to learn efficiently on a moment-by-moment basis using memory and computational complexity that is constant in time. These methods are also distinguished due to their ability to learn from other predictions, a technique known as \emph{bootstrapping}, which often provides fast and more accurate answers (Sutton 1988).


TD algorithms conventionally make updates at every state visited,
implicitly giving higher importance, in terms of function-approximation resources, 
to states that are visited more frequently. 
As the value cannot be estimated accurately under function approximation, 
valuing some states more means valuing others less. 
We may, however, be interested in valuing some states more than others 
based on criteria other than visitation frequency.
Conventional TD updates do not provide that flexibility and 
cannot be naively modified. 
For example, in the case of off-policy TD updates, updating according to one policy while learning about another can cause divergence (Baird 1995).

In this paper, we discuss emphatic TD\la (Sutton et al.\ 2015), a principled solution for the problem of selective updating, where convergence is ensured under an arbitrary interest in visited states as well as off-policy training. The idea is to emphasize and de-emphasize state updates with user-specific interest in conjunction with how much other states bootstrap from that state. 
We first describe this idea in a simpler case: linear function approximation with full bootstrapping (i.e., $\lambda=0$). We then derive the full algorithm for the more general off-policy learning setting with arbitrary bootstrapping. Finally, after briefly summarizing the available results on the stability and convergence of the new algorithm, we discuss the use and the potential advantages of this algorithm using an illustrative experiment.

\section{The problem of selective updates}

Let us start with the problem of selective updating in the
simplest function approximation case: linear TD\la with $\l=0$.
Consider a Markov decision process (MDP) with a finite set $\SS$ of $N$ states and a finite set $\AA$ of actions, for the discounted total reward criterion with discount rate $\g\in[0,1)$. In this setting, an agent interacts with the environment by taking an action $A_t\in\AA$ at state $S_t \in \SS$ according to a policy $\pi:\AA\times\SS\ra[0,1]$ where $\pi(a|s) \doteq \CP{A_t\!=\!a}{S_t\!=\!s}$\footnote{The notation $\doteq$ indicates an equality by definition.},
transitions to state $S_{t+1}\in\SS$, and receives reward $R_{t+1}\in\Re$ in a sequence of time steps $t \geq 0$. Let $\Ppi \in \Re^{N\times N}$ denote the state transition probability matrix and $\rr_\pi\in\Re^N$ the expected immediate rewards from each state under $\pi$. The value of a state is then defined as:
\begin{align}
\vpi(s) \doteq \CEpi{G_t}{S_t\!=\!s}, \label{eq:vpi}
\end{align}
where $\Epi{\cdot}$ denotes an expectation conditional on all actions being selected according to $\pi$, and $G_t$, the \emph{return} at time $t$, is a random variable of the future outcome:
\begin{align}
G_t &\doteq R_{t+1} + \g R_{t+2} + \g^2 R_{t+3} + \cdots. \label{eq:G}
\end{align}
We approximate the value of a state as a linear function of its features: $\th\tr\ph(s) \approx \vpi(s)$, where $\ph(s)\in\Re^n$ is the feature vector corresponding to state $s$. 
Conventional linear TD(0) learns the value function $\vpi$ by generating a sequence of parameter vectors $\th_t\in\Re^n$:
\begin{align} \label{eq:TD0}
  \th_{t+1} \doteq \th_t + \a\left(R_{t+1}+\g\th_t\tr\ph(S_{t+1})-\th_t\tr\ph(S_t)\right)\ph(S_t), 
\end{align}
where $\a>0$ is a step-size parameter.

Additionally, we may have a relative interest in each state, denoted
by a nonnegative \emph{interest function} $i:\SS\ra[0,\infty)$. For example, in episodic problems we often care primarily about the value of the first state, or of earlier states generally (Thomas 2014). 
A straightforward way to incorporate the relative interests into TD(0) would be to use $i(S_t)$ as a factor to the update on each state $S_t$:
\begin{align} 
  \th_{t+1} \doteq \th_t + \a \left(R_{t+1}+\g\th_t\tr\ph(S_{t+1})-\th_t\tr\ph(S_t)\right)i(S_t)\ph(S_t).
\end{align}
 
In order to illustrate the problem of this approach, suppose there is a Markov chain consisting of two non-terminal and a terminal state with features $\ph(1) = 1$ and $\ph(2) = 2$ and interests $i(1) = 1$ and $i(2) = 0$ (cf. Tsitsiklis \& Van Roy 1996):

\vspace{10pt}
\centerline{\includegraphics[height=0.3in]{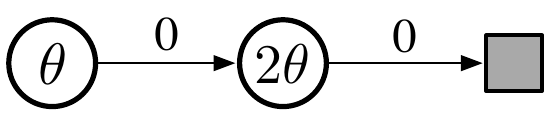}}
\vspace*{1pt}

\noindent
Then the estimated values are $\thet$ and $2\thet$ for a scalar parameter $\thet \in \mathbb{R}$. Suppose that $\thet$ is 10, the reward on the first transition is 0. The transition is then from a state valued at 10 to a state valued at 20. If $\g=1$ and $\a$ is 0.1, then $\thet$ will be increased to $11$. But then the next time the transition occurs there will be an even bigger increase in value, from 11 to 22, and a bigger increase in $\thet$, to $12.1$. If this transition is experienced repeatedly on its own, then the system is unstable and the parameter increases without bound---it diverges.

This problem arises due to both bootstrapping and the use of function approximation, which entails shared resources among the states. If a tabular representation was used instead, the value of each state would be stored independently and divergence would not occur. Likewise, if the value estimate of the first state was updated without bootstrapping from that of the second state, such divergence could again be avoided. 

\emph{Emphatic TD(0)} (Sutton et al.\ 2015) remedies this problem of TD(0) by emphasizing the update of a state, depending on how much a state is bootstrapped in conjunction with the relative interest in that state. Although $\l=0$ gives full bootstrapping, the amount of bootstrapping is still modulated by $\g$. For example, if $\g=0$, then no bootstrapping occurs even with $\l=0$. 
The amount of emphasis to the update of a state at time $t$ is:
\begin{align}
F_t	&\doteq i(S_t) + \g i(S_{t-1}) + \g^2 i(S_{t-2}) + \cdots + \g^t i(S_0) = i(S_t) + \g F_{t-1}.
\end{align}
The following update defines emphatic TD(0):
\begin{align*}
\th_{t+1} 	&\doteq \th_t + \a \left(R_{t+1}+\g\th_t\tr\ph(S_{t+1})-\th_t\tr\ph(S_t)\right)F_t\ph(S_t).
\end{align*}
According to this algorithm, the value estimate of a state is updated if the user is interested in that state or it is reachable from another state in which the user is interested. 
Going back to the above two-state example, 
the second state value is now also updated despite having a user-specified interest of 0.
In fact, $F_t$ is equal for both states, 
and updating is exactly equivalent to on-policy sampling; hence, divergence does not occur. 
For other choices of relative interest and discount rate, the effective state distribution can be different than on-policy sampling, but the algorithm still converges as we show later.

\section{\ETD\la: The off-policy emphatic TD\la}

In this section, we develop the emphatic TD algorithm, which we call \emph{\ETD\la,}  in the generic setting of off-policy training with state-dependent discounting and bootstrapping. 

Let $\g:\SS\ra[0,1]$ be the state-dependent degree of discounting; equivalently, $1-\g(s)$ is the probability of terminating upon arrival in state $s$. 
Let $\l:\SS\ra[0,1]$ denote a state-dependent degree of bootstrapping; 
in particular, $1-\l(s)$ determines the degree of bootstrapping upon arriving in state $s$. 
As notational shorthand, we use  $\g_t\doteq \g(S_t)$, $\l_t \doteq \l(S_t)$, and $\ph_t\doteq \ph(S_t)$.
For TD learning, we define a general notion of bootstrapped return, the \emph{$\l$-return}, with state-dependent bootstrapping and discounting, by
\begin{align*}
G_t^\l	\doteq R_{t+1} + \g_{t+1} \lt( (1-\l_{t+1})\th_t\tr\ph_{t+1} + \l_{t+1} G_{t+1}^\l \rt).
\end{align*}
This return can be directly used to estimate $v_\pi$ on-policy as long as the agent follows $\pi$. 
However, in off-policy learning, experience is generated by following a different policy $\mu:\AA\times\SS\ra[0,1]$, often called the \emph{behavior} policy. 
To obtain an unbiased estimate of the return under $\pi$,
the experience generated under $\mu$ has to be reweighted by 
importance sampling ratios: $\r_t \doteq \frac{ \pi(A_t|S_t) }{\mu(A_t|S_t)}$,
assuming $\mu(a|s) > 0$ for every state and action for which $\pi(a|s) > 0$.
The importance-sampled $\l$-return for off-policy learning is thus defined as follows (Maei 2011, van Hasselt et al. 2014):
\begin{align*}
G_t^{\l\r}	\doteq \r_t\lt(R_{t+1} + \g_{t+1} \lt( (1-\l_{t+1}\r_{t+1})\th_t\tr\ph_{t+1} + \l_{t+1} G_{t+1}^{\l\r} \rt)\rt).
\end{align*}
\noindent

The forward-view update of the conventional off-policy TD\la can be written as:
\begin{align}
\th_{t+1}		\doteq \th_t + \a \lt(G_t^{\l\r} - \r_t\ph_t\tr\th_t \rt) \ph_t. \label{eq:off-td-for}
\end{align}
The backward-view update with an offline equivalence (cf.\ van Seijen \& Sutton 2014) with the above forward view can be written as:
\begin{align}
 \th_{t+1} &\doteq  \th_t + \a\left(R_{t+1} + \g_{t+1}\th_t\tr\ph_{t+1} - \th_t\tr\ph_t\right)\ee_t  \label{eq:tht}\\
     \ee_t &\doteq  \rho_t\left(\g_t\l_t\ee_{t-1} + \ph_t \right),\text{~~~~with~}\ee_{-1}\doteq \bm 0\label{eq:et},
\end{align}
where $\ee_t\in\Re^n$ is the eligibility-trace vector at time $t$. This algorithm makes an update to each state visited under $\mu$ and does not allow user-specified relative interests to different states. Convergence is also not guaranteed in general for this update rule.

By contrast, instead of (\ref{eq:off-td-for}), we define the forward view of \ETD\la to be:
\begin{align}
\th_{t+1} 	&\doteq \th_t + \a \lt(G_t^{\l\r} - \r_t\ph_t\tr\th_t \rt) M_t \ph_t. \label{eq:etd-for}
\end{align}
Here $M_t\in\Re$ denotes the emphasis given to update at time $t$, and it is derived based on the following reasoning, similar to the derivation of $F_t$ for emphatic TD(0).

The emphasis to the update at state $S_t$ is first and foremost, due to $i(S_t)$, the inherent interest of the user to that state. A portion of the emphasis is also due to the amount of bootstrapping the preceding state $S_{t-1}$ does from $S_t$, determined by $\g_t(1-\l_t)\r_{t-1}$: the probability of not terminating at $S_t$ times the probability of bootstrapping at $S_t$  times the degree by which the preceding transition is followed under the target policy. Finally, $M_t$ also depends on $M_{t-1}$, the emphasis of the preceding state itself. 
The emphasis for state $S_t$ similarly depends on all the preceding states that bootstrap from this state to some extent. Thus the total emphasis can be written as:
\begin{align}
M_t 	&\doteq i(S_t) + \sum_{k=0}^{t-1} M_k \r_k \lt( \prod_{i=k+1}^{t-1}\g_i\l_i\r_i \rt) \g_t(1-\l_t) = \l_t i(S_t) + (1-\l_t) F_t, \label{eq:Mt}
\end{align}

\vspace{.3cm}
\noindent where
\vspace{-.95cm}
\begin{flalign}
\quad F_{t} 
\doteq i(S_t) + \g_t\sum_{k=0}^{t-1} \rho_k M_k \prod_{i=k+1}^{t-1} \g_i \l_i \rho_i = i(S_t) + \g_{t} \rho_{t-1}F_{t-1},\ \ \text{with~}F_{-1}\doteq 0, \label{eq:Ft}
\end{flalign}
\noindent
giving the final update for \ETD\la, derived from the forward-view update (\ref{eq:etd-for}):
\begin{align}
 \th_{t+1} &\doteq  \th_t + \a\left(R_{t+1} + \g_{t+1}\th_t\tr\ph_{t+1} - \th_t\tr\ph_t\right)\ee_t \label{eq:etd} \\
     \ee_t &\doteq  \rho_t\left(\g_t\l_t\ee_{t-1} + M_t\ph_t \right),\text{~~~~with~}\ee_{-1}\doteq \bm 0 
     .\label{eq:et2}
\end{align}
The trace $F_t$ here is similar to that of emphatic TD(0), adapted to the off-policy case through the application of $\r_t$. According to (\ref{eq:Mt}), the emphasis $M_t$ can be written simply as a linear interpolation between $i(S_t)$ and $F_t$. The per-step computational and memory complexity of \ETD\la is the same as that of original TD\la: $O(n)$ in the number of features. The additional cost \ETD\la incurs due to the computation of the scalar emphasis is negligible.

\section{Stability and convergence of \ETD\la}
We have discussed the motivations and ideas that led to the design of the emphasis weighting scheme (\ref{eq:Mt})-(\ref{eq:et2}) for \ETD\la. We now discuss several salient analytical properties underlying the algorithm due to this weighting scheme, and present the key stability and convergence results we have obtained for the algorithm.\ First, we formally state the conditions needed for the analysis.\vspace*{-0.23cm}

\begin{assumption}[Conditions on the target and behavior policies] \label{cond:policies} \hfill\vspace*{-0.2cm}
\begin{itemize}
\item[(i)] The target policy $\pi$ is such that $(\I - \Ppi \Gm)^{-1}$ exists, where $\Gm$ is the $N \times N$ diagonal matrix with the state-dependent discount factors $\g(s), s \in \SS$, as its diagonal entries.\vspace*{-0.25cm} 
\item[(ii)] The behavior policy $\mu$ induces an irreducible Markov chain on $\SS$, with the unique invariant distribution $d_\mu(s), s \in \SS$, and for all $(s,a) \in \SS \times \AA$, $\mu(a | s) > 0$ if $\pi(a | s) > 0$.\vspace*{-0.1cm}
\end{itemize}
\end{assumption}

Under Assumption~\ref{cond:policies}(i), the value function $v_\pi$ is specified by the expected total (discounted) rewards as $\Vpi = (\I - \Ppi \, \Gm)^{-1} \rr_\pi$; i.e., $\Vpi$ is the unique solution of the Bellman equation
$\vv = \rr_\pi + \Ppi \, \Gm \vv.$
Associated with \ETD\la is a multistep, generalized Bellman equation which is determined by the bootstrapping parameters $\l(s)$ and also has $\Vpi$ as its unique solution (Sutton 1995):
\begin{equation} 
   \vv = \rr_\pi^\l + \Ppi^\l  \vv, \label{eq:Bellman}
\end{equation}   
where $\Ppi^\l$ is a substochastic matrix and $\rr_\pi^\l \in \Re^N$.
\footnote{Specifically, with $\Lm$ denoting the diagonal matrix with $\l(s), s \in \SS$, as its diagonal entries, we have
$\Ppi^\l =  \I - (\I - \Ppi \Gm \Lm)^{-1} \, (\I - \Ppi \Gm)$ and $\rr_\pi^\l = (\I - \Ppi \Gm \Lm)^{-1} \, \rr_\pi$.} 
Let $\Ph$ be the $N \times n$ matrix with the feature vectors $\ph(s)\tr, s \in \SS$, as its rows. The goal of \ETD\la is to find an approximate solution of the Bellman equation (\ref{eq:Bellman}) in the space $\{ \Ph \th \mid \th \in \Re^n \}$.

Let us call those states on which \ETD\la places positive emphasis weights \emph{emphasized states}. More precisely, under Assumption~\ref{cond:policies}(ii), we can assign an expected emphasis weight $\m(s)$ for each state $s$, according to the weighting scheme (\ref{eq:Mt})-(\ref{eq:et2}), as (Sutton et al.\ 2015):
\begin{equation}
 \big[ \, \m(1), \, \m(2), \, \ldots, \, \m(N) \, \big] = \Dmi\tr (\I - \Ppi^\l)^{-1}, \label{eq:emphasis}
 \end{equation}
where $\Dmi \in \Re^N$ denotes the vector with components $\dmi(s) = \dm(s) \cdot i(s), s \in \SS$. Emphasized states are precisely those with $\m(s) > 0$. It is important to observe from (\ref{eq:emphasis}) that the emphasis weights $\m(s)$ reflect the occupancy probabilities of the \emph{target policy}, with respect to $\Ppi^\l$ and an initial distribution proportional to $\Dmi$, rather than the behavior policy. As will be seen shortly, this gives \ETD\la a desired stability property that lacks normally in TD\la algorithms with selective updating.

Let $\M$ denote the diagonal matrix with the emphasis weights $\m(s)$ on its diagonal.
By considering the stationary case, the equation that \ETD\la aims to solve is shown by Sutton et al.\ (2015) to be:
\begin{equation} 
     \A \th = \b, \quad \th \in \Re^n, \label{eq:ETDeq-th}
\end{equation}
\begin{equation} 
\text{where} \qquad \A  = \Ph\tr  \M \, (\I - \Ppi^\l) \,  \Ph , \qquad 
    \b   =  \Ph\tr  \M \,  \rr_\pi^\l. \qquad \qquad \qquad
\end{equation}      
In terms of the approximate value function $\vv = \Ph \th$, under a mild condition on the approximation architecture given below, the equation (\ref{eq:ETDeq-th}) is equivalent to a projected version of the Bellman equation (\ref{eq:Bellman}):
\begin{equation}
  \vv = \Pi \big(\rr_\pi^\l + \Ppi^\l  \vv \big), \quad \vv \in \big\{ \Ph \th \mid \th \in \Re^n\big\},\label{eq:ETDeq-v}
\end{equation}  
where $\Pi$ denotes projection onto the approximation subspace with respect to a weighted Euclidean norm or seminorm $\| \cdot \|^2_m$, defined by the emphasis weights as $\| \vv\|^2_m = \sum_{s \in \SS} \m(s) v(s)^2$.

\begin{assumption}[Condition on the approximation architecture] \label{cond:features} \hfill \\
The set of feature vectors of emphasized states, $\{\ph(s) \mid s \in \SS, \, \m(s) > 0 \}$, contains $n$ linearly independent vectors.
\end{assumption}
\vspace*{-0.1cm}

We note that Assumption~\ref{cond:features} (which implies the linear independence of the columns of $\Phi$) is satisfied in particular if the set of feature vectors, $\{\ph(s) \mid s \in \SS, \, i(s) > 0 \}$, contains $n$ linearly independent vectors, since states with positive interest $i(s)$ are among the emphasized states. So this assumption can be easily satisfied in reinforcement learning without model knowledge.

We are now ready to discuss an important stability property underlying our algorithm. 
By making the emphasis weights $\m(s)$ reflecting the occupancy probabilities of the target policy, as discussed earlier, the weighting scheme (\ref{eq:Mt})-(\ref{eq:et2}) of our algorithm ensures that the matrix $\A$ is positive definite under almost minimal conditions for off-policy training:
\footnote{The conclusion of Theorem~\ref{thm:matrix} for the case of an interest function $i(\cdot) > 0$ is first proved by Sutton, Mahmood, and White (see their Theorem 1); Theorem~\ref{thm:matrix} as given here is proved by Yu (2015) (see Prop.\ C.2 and Remark C.2 in Appendix C therein). The analyses in both works are motivated by a proof idea of Sutton (1988), which is to analyze the structure of the $N \times N$ matrix $\M (\I - \Ppi^\l)$ and to invoke a result from matrix theory on strictly or irreducibly diagonally dominant matrices (Varga 2000, Cor.\ 1.22).}

\vspace*{-0.0cm}
\begin{theorem}[Stability property of $\A$] \label{thm:matrix}
Under Assumptions \ref{cond:policies}-\ref{cond:features}, the matrix $\A$ is positive definite (that is, there exists $c > 0$ such that $\th\tr \A \th \geq c \, \| \th \|_2^2$ for all $\th \in \Re^n$). 
\end{theorem}

This property of $\A$ shows that the equation (\ref{eq:ETDeq-th}) associated with \ETD\la has a unique solution $\th^*$ (equivalently, the equation (\ref{eq:ETDeq-v}) has the approximate value function $\vv = \Ph \th^*$ as its unique solution). Moreover, it shows that unlike normal TD\la with selective updating, here the deterministic update in the parameter space, $\th_{t+1} = \th_t - \a (\A \th_t - \b)$, converges to $\th^*$ for sufficiently small stepsize $\a$, and when diminishing stepsizes $\{\alpha_t\}$ are used in \ETD\la, $\{\th^*\}$ is globally asymptotically stable for the associated ``mean ODE'' $\dot{\th} = - \A \th + \b$ (Kushner \& Yin 2003).
\footnote{The important analytical properties discussed here can be shown to also extend to the case where the linear independence condition in Assumption~\ref{cond:features} is relaxed: there, $\A$ acts like a positive definite matrix on the subspace of $\th$ (the range space of $\A$) that \ETD\la naturally operates on. These extensions are based on both our understanding of how the weighting scheme (\ref{eq:Mt})-(\ref{eq:et2}) is designed (Sutton et al.\ 2015) and the special structure of the matrix $\M(\I - \Ppi^\l)$ revealed in the proof of (Yu 2015, Prop.\ C.2). We will report the details of these extensions in a separate paper, however.}
We are now ready to address the convergence of the algorithm.

\vspace*{-0.1cm}
\begin{assumption}[Conditions on noisy rewards and diminishing stepsizes] \label{cond:noise-stepsize} \hfill\vspace*{-0.15cm}
\begin{itemize}
\item[(i)] The variances of the random rewards $\{R_t\}$ are bounded.\vspace*{-0.2cm}
\item[(ii)] The (deterministic) stepsizes $\{\a_t\}$ satisfy that $\a_t = O(1/t)$ and $\frac{\a_t - \a_{t+1}}{a_t} = O(1/t)$.
\end{itemize}
\end{assumption}

Under the preceding assumptions, we have the following result, proved in (Yu 2015):
\footnote{The proof is similar to but more complex than the convergence proof for off-policy LSTD/TD (Yu 2012). Among others, we show that despite the high variance in off-policy learning, the Markov chain $\{(S_t, A_t, \ee_t, F_t)\}$ on the joint space $\SS \times \AA \times \Re^{n+1}$ exhibits nice properties including ergodicity. We use these properties together with convergence results for a least-squares version of \ETD\la and a convergence theorem from stochastic approximation theory (Kushner \& Yin 2003, Theorem 6.1.1) to establish the desired convergence of \ETD\la and its constrained variant by a ``mean ODE'' based proof method.
}

\begin{theorem}[Convergence of \ETD\la] \label{thm:alg} \hfill \\
Let Assumptions \ref{cond:policies}-\ref{cond:noise-stepsize} hold. Then, for each initial $\th_0 \in \Re^n$, the sequence $\{\th_t\}$ generated by \ETD\la converges to $\th^*$ with probability $1$.
\end{theorem}

To satisfy the stepsize Assumption~\ref{cond:noise-stepsize}(ii), we can take $\a_t =  c_1/(c_2+t)$ for some constants $c_1,c_2 > 0$, for example. If the behavior policy is close to the target policy, we believe that \ETD\la also converges for larger stepsizes.

\section{An illustrative experiment}

\def\behProb{0.4}
\def\wisProb{0.6}

\begin{wrapfigure}{L}{2.4in}
\vspace{-.5cm}
{\includegraphics[height=2.3in]{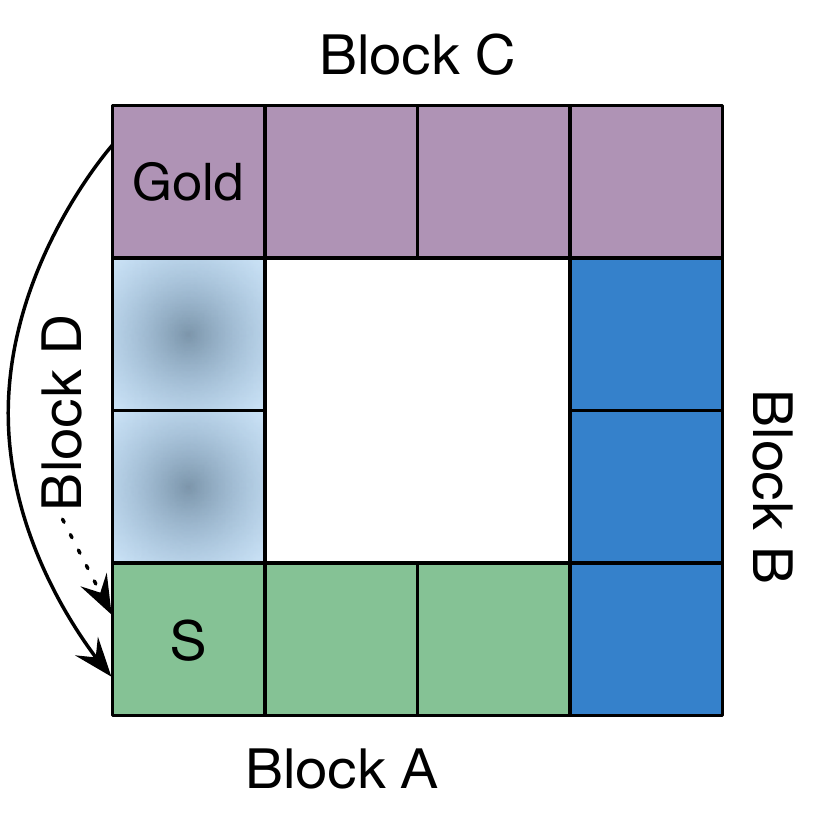}}

{Figure 1. The \emph{Miner} problem where a miner continually collects gold from the {\ttfamily Gold} cell until it falls into a trap, which can be activated in \mbox{{\ttfamily Block D}}.}
\end{wrapfigure}

In this section we describe an experiment to illustrate the flexibility and benefits of \ETD\la in learning several off-policy predictions in terms of value estimates. 

In this experiment we used a gridworld problem depicted in Figure 1, which we call the \emph{Miner} problem. Here a miner starting from the cell {\ttfamily S} continually wandered around the gridworld using one of the following actions: {\ttfamily left}, {\ttfamily right}, {\ttfamily up} and {\ttfamily down}, each indicating the direction of the miner's movement. An invalid direction such as going down from {\ttfamily S} resulted in no movement. The miner got zero reward at every transition except when it arrived at the cell denoted by {\ttfamily Gold}, in which case a $+1$ reward was obtained. There were two routes to reach the {\ttfamily Gold} cell from {\ttfamily S}: one went straight up through {\ttfamily Block D}, and the other was roundabout through {\ttfamily Block B}. A trap could be activated in one of the two cells in {\ttfamily Block D} chosen randomly. Once active, a trap stayed for 3 time steps, and only one trap was active at any time. The trap activation probability was 0.25. If the miner arrived at the {\ttfamily Gold} cell or fell into a trap, it was transported to {\ttfamily S} in the next time step. Note that arriving at the {\ttfamily Gold} cell or a trap was not the end of an episode, and the miner wandered around continually.

The miner followed a fixed behavior policy according to which the miner was equally likely to take any of the four actions in {\ttfamily Block A}, more inclined to go up in both {\ttfamily Block B} and {\ttfamily Block D}, and more inclined to go left in {\ttfamily Block C}, in each case with probability \behProb. The rest of the actions were equally likely.

We evaluated three fixed policies different than the behavior policy. We call them {\ttfamily uniform}, {\ttfamily headfirst} and {\ttfamily cautious} policies. Under the {\ttfamily uniform} policy, all actions were equally likely in every cell. Under the {\ttfamily headfirst} policy, the miner chose to go up in {\ttfamily Block A} and {\ttfamily D} with 0.9 probability while other actions from those blocks were equally likely. All the actions from other blocks were chosen with equal probability. Under the {\ttfamily cautious} policy, the miner was more inclined to go right in {\ttfamily Block A}, go up in both {\ttfamily Block B} and {\ttfamily Block D}, and go left in {\ttfamily Block C}, in each case with probability \wisProb. The rest of the actions were equally likely.  

We were interested to predict how much gold the miner could collect before falling into a trap if the miner had used the above three policies, without executing any of these policies. We set $\g=0$ for those states where the miner got entrapped to indicate termination under the target policy (although behavior policy continued) and a discounting of $\g=0.99$ in other states. We set $i(s)=1$ whenever the miner was in {\ttfamily Block A} and 0 everywhere else. As the behavior policy of the miner is different than the three target policies, it must use off-policy training to learn what could happen under each of those policies. We used three instances of ETD\la for three different predictions, each using $\a=0.001$, $\l=1.0$ when the miner was in {\ttfamily Block D}, $\l=0$ in {\ttfamily Block A}, and $\l=0.9$ in other states. We clipped each component of the increment to $\th$ in (\ref{eq:etd}) between $-0.5$ and $+0.5$ in order to reduce the impact of extremely large eligibility traces on updates. Clipping the increments can be shown to be theoretically sound, although we will not discuss this subject here. The state representation used four features: each corresponding to the miner being in one of the four blocks. The miner wandered continually until $3000$ entrapments occurred.

\begin{wrapfigure}{L}{2.5in}
\vspace{-.5cm}
{\includegraphics[width=2.5in]{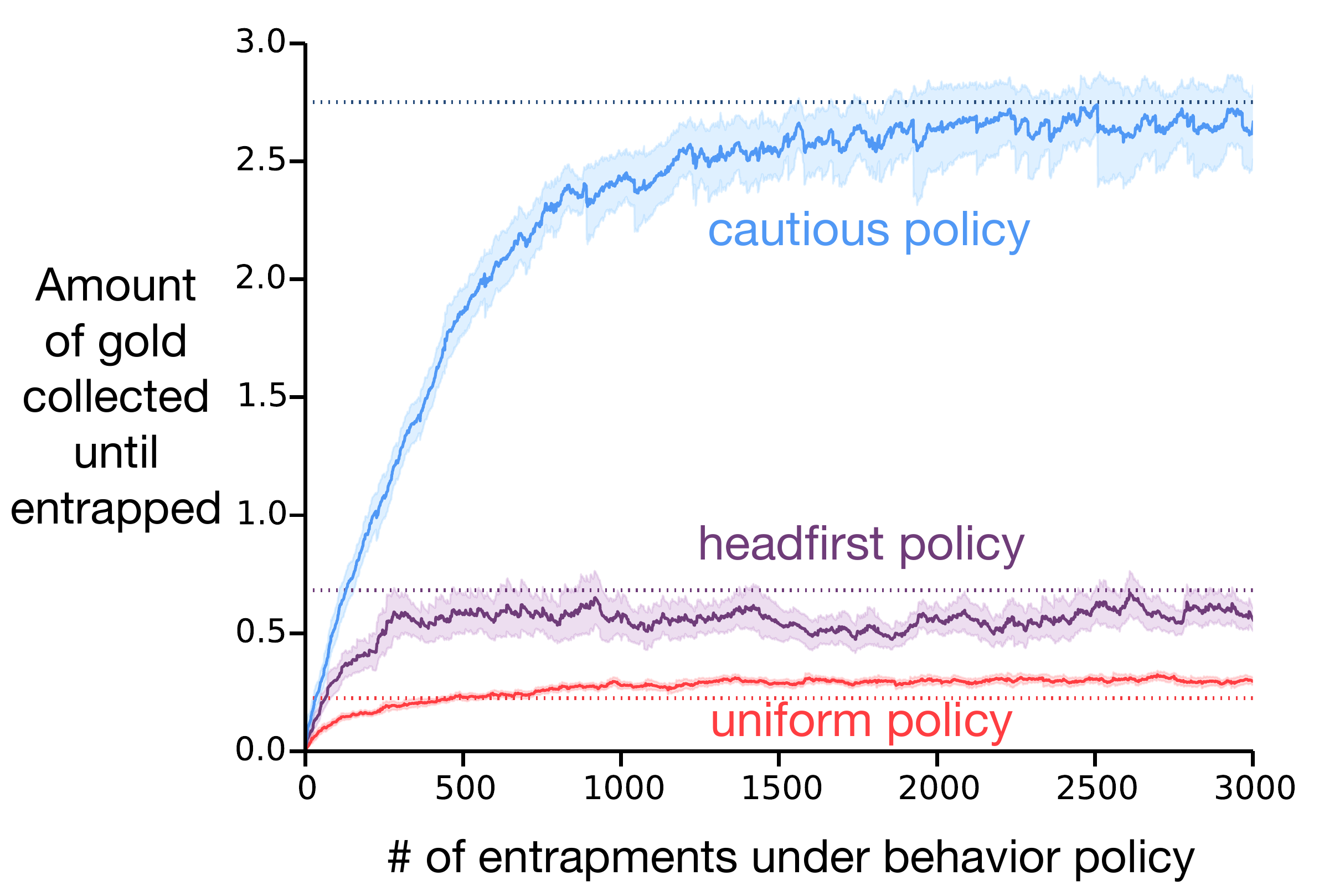}}
{Figure 2. 
Simultaneous evaluation of three policies different than the behavior policy using ETD\la.
}
\end{wrapfigure}
Figure 2 shows estimates calculated by ETD\la in terms of its weight corresponding to {\ttfamily Block A} for the three target policies. The curves shown are average estimates with two standard error bands using $50$ independent runs. The dotted straight lines indicate the true state value estimated through Monte Carlo simulation from {\ttfamily S}. 
Due to the use of function approximation and clipping of the updates, the true value could not be estimated accurately. However, the estimates for the three policies appear to approach values close to the true ones, and they preserved the relative ordering of the policies. In the absence of the clipping, the estimates were less stable and highly volatile, occasionally moving far away from the desired value for some of the runs. Although some of the learning curves still look volatile, clipping the updates reduced its extent considerably.


\vspace{-.2cm}
\section{Discussion and conclusions}
We summarized the motivations, key ideas and the available results on emphatic algorithms. Furthermore, we demonstrated how \ETD\la can be used to learn many predictions about the world simultaneously using off-policy learning, and the flexibility it provides through state-dependent discounting, bootstrapping and user-specified relative interests to states. \ETD\la is among the few algorithms with per-step linear computational complexity that are convergent under off-policy training. Compared to convergent gradient-based TD algorithms (Maei 2011), \ETD\la is simpler and easier to use; it has only one learned parameter vector and one step-size parameter. The problem of high variance is common in off-policy learning, and \ETD\la is susceptible to it as well. An extension to variance-reduction methods, such as weighted importance sampling (Precup et al.\ 2000, Mahmood et al.\ 2014, 2015), can be a natural remedy to this problem. \ETD\la produces a different algorithm than the conventional TD\la even in the on-policy case. It is likely that, in many cases, \ETD\la provides more accurate predictions than TD\la through the use of relative interests and emphasis. An interesting direction for future work would be to characterize these cases.


{
\section*{References}
\footnotesize
\parindent=0pt
\def\hangin{\hangindent=0.15in}
\parskip=6pt

\hangin
Baird, L.~C. (1995).
\newblock Residual algorithms: {R}einforcement learning with function
  approximation.
\newblock In {\em Proceedings of the
 12th International Conference on Machine Learning}, pp.~30--37.

\hangin
Bertsekas, D.~P., Tsitsiklis, J.~N. (1996).
\newblock {\em Neuro-Dynamic Programming}.
\newblock Athena Scientific. 

\hangin
Boyan, J.~A., (1999).
Least-squares temporal difference learning.
In \emph{Proceedings of the 16th International Conference on Machine Learning}, pp. 49--56. 

\hangin
Bradtke, S., Barto, A.~G. (1996).
\newblock Linear least-squares algorithms for temporal difference learning.
\newblock {\em Machine Learning 22}:33--57.


\hangin
Kushner, H.~J., Yin G.~G. (2003).
\emph{Stochastic Approximation and Recursive Algorithms and Applications}, second edition.
Springer-Verlag.

\hangin
Maei, H.~R., Sutton, R.~S. (2010). 
GQ($\l$): A general gradient algorithm for temporal-difference prediction learning with eligibility traces. 
In  \emph{Proceedings of the Third Conference on Artificial General Intelligence}, pp.~91--96. Atlantis Press.

\hangin
Maei, H.~R. (2011).
\emph{Gradient Temporal-Difference Learning Algorithms}. 
PhD thesis, University of Alberta. 

\hangin
Mahmood, A.~R., van Hasselt, H., Sutton, R.~S. (2014).
Weighted importance sampling for off-policy learning with linear function approximation.
\textit{Advances in Neural Information Processing Systems 27}.

\hangin
Mahmood, A. R., Sutton, R. S. (2015).
Off-policy learning based on weighted importance sampling with linear computational complexity. In \emph{Proceedings of the 31st Conference on Uncertainty in Artificial Intelligence}, Amsterdam, Netherlands.

\hangin
Precup, D., Sutton, R.~S., Singh, S. (2000).
Eligibility traces for off-policy policy evaluation. 
In \emph{Proceedings of the 17th International Conference on Machine Learning}, 
pp.~759--766. Morgan Kaufmann.

\hangin
Sutton R.~S. (1988).
Learning to predict by the methods of temporal differences.
{\em Machine Learning} 3:9-44.
  
\hangin
Sutton R.~S. (1995).
TD models: Modeling the world at a mixture of time scales.
In \emph{Proceedings of the 12th International Conference on Machine Learning}.

\hangin
Sutton, R.~S., Barto, A.~G. (1998). \emph{Reinforcement Learning: An Introduction}. MIT Press.


\hangin
Sutton, R. S., Mahmood, A.~R., White, M.\ (2015). An emphatic approach to the problem of off-policy temporal-difference learning, arXiv:1503.04269 [cs.LG].

\hangin
Thomas, P. (2014).
Bias in natural actor--critic algorithms.
In \emph{Proceedings of the 31st International Conference on Machine Learning}.
JMLR W\&CP 32(1):441--448.

\hangin
Tsitsiklis, J.~N., {Van Roy}, B. (1996).
\newblock Feature-based methods for large scale dynamic programming.
\newblock {\em Machine Learning} 22:59--94.


\hangin
Varga, R.~S. (2000). \emph{Matrix Iterative Analysis},  second edition. Springer-Verlag.  

\vspace{.1cm}
\hangin
van Hasselt, H., Mahmood, A.~R., Sutton, R.~S. (2014). Off-policy TD\la with a true online equivalence. In \emph{Proceedings of the 30th Conference on Uncertainty in Artificial Intelligence}, Quebec City, Canada.

\hangin
van Seijen, H., \& Sutton, R.~S. (2014).
True online TD($\l$).
In \emph{Proceedings of the 31st International Conference on Machine Learning}.
JMLR W\&CP 32(1):692--700.

\hangin
Yu, H. (2012). 
Least squares temporal difference methods: An analysis under general conditions.
{\em SIAM Journal on Control and Optimization} 50:3310-3343.

\hangin
Yu, H. (2015). On convergence of emphatic temporal-difference learning. In \emph{Proceedings of  the 28th Annual Conference on Learning Theory}. Paris, France.
}

\end{document}